\RequirePackage{snapshot}
\documentclass[runningheads]{llncs}
\usepackage{graphicx}
\usepackage{amsmath,amssymb} 
\usepackage{color}
\usepackage[width=122mm,left=12mm,paperwidth=146mm,height=193mm,top=12mm,paperheight=217mm]{geometry}
\usepackage{xr}

\usepackage[utf8]{inputenc}
\graphicspath{{figures/}}

\usepackage[font=small,skip=5pt]{caption}
\usepackage{tikz}
\usepackage{tkz-euclide}
\usetikzlibrary{calc}
\usetikzlibrary{math}
\usetikzlibrary{shapes}
\usetikzlibrary{decorations.pathreplacing}
\usepackage{dsfont}
\usepackage{diagbox}
\usepackage{multirow}
\usepackage{csvsimple}
\usepackage[normalem]{ulem}
\usepackage{relsize}
\newcommand\noktikz{}

\DeclareMathOperator*{\argmax}{arg\,max}
\DeclareMathOperator*{\argmin}{arg\,min}

\newcommand{\fix}{\text{fix}}
\newcommand{\nonfix}{\text{nonfix}}
\newcommand{\EE}{\mathds{E}}

\newcommand{\myparagraph}[1]{\vspace{-0.3cm}\paragraph{\textbf{#1}}}

\begin{document}
    
\pagestyle{headings}
\mainmatter
\def\ECCV18SubNumber{1846}  

\title{Saliency Benchmarking Made Easy:\\ Separating Models, Maps and Metrics}

\titlerunning{Saliency Benchmarking Made Easy}

\author{Matthias Kümmerer\inst{1}
\and Thomas S.A. Wallis\inst{1,2}
\and Matthias Bethge\inst{1}}

\institute{Werner Reichardt Centre for Integrative Neuroscience, University of Tübingen, Tübingen, Germany
 \and
Wilhelm-Schickard Institute for Computer Science (Informatik), University of Tübingen, Tübingen, Germany\\
\email{\{matthias.kuemmerer,tom.wallis,matthias\}@bethgelab.org}}

\authorrunning{M. Kümmerer, T.S.A. Wallis and M. Bethge}

\maketitle

\begin{abstract}
    Dozens of new models on fixation prediction are published every year and compared on open benchmarks such as MIT300 and LSUN.
    However, progress in the field can be difficult to judge because models are compared using a variety of inconsistent metrics.
    Here we show that no single saliency map can perform well under all metrics.
    Instead, we propose a principled approach to solve the benchmarking problem by separating the notions of saliency models, maps and metrics.
Inspired by Bayesian decision theory, we define a saliency model to be a probabilistic model of fixation density prediction and a saliency map to be a metric-specific prediction derived from the model density which maximizes the expected performance on that metric given the model density.
We derive these optimal saliency maps for the most commonly used saliency metrics (AUC, sAUC, NSS, CC, SIM, KL-Div) and show that they can be computed analytically or approximated with high precision.
We show that this leads to consistent rankings in all metrics and avoids the penalties of using one saliency map for all metrics.
Our method allows researchers to have their model compete on many different metrics with state-of-the-art in those metrics:
``good'' models will perform well in all metrics.

\end{abstract}

\section{Introduction}





Humans have a foveated visual system: only a small central part of the retina has high receptor density allowing the perception of the details of a scene. 
Therefore humans make eye movements to place the high resolution fovea on things they want to see. 
Understanding where they choose to look is therefore an important component of understanding behaviour.


A long-standing account of bottom-up attentional guidance posits the existence of a ``saliency map'' (or maps) in the human brain \cite{treisman1980feature,koch_shifts_1985}.
Here, a saliency map represents spatial importance, usually defined to be local contrast in low-level features such as luminance, color or orientation. 
Since Itti and Koch formulated this concept into their seminal image-based model \cite{IttiKoch1998Model}, a large number of models have been proposed for predicting fixations from image features, e.g. \cite{Harel2006GBVS,Zhang2008SUN,kienzle2009,BruceTsotso2009Saliency,Judd2009Model,zhang2013saliency,adeli2017colliculus} and more recently many models based on deep learning, e.g. \cite{Vig2014,kuemmerer2015deepgaze,Huang_2015_ICCV,Kruthiventi2015,pan2017salgan,kuemmerer2017iccv}; see \cite{borji_state_of_the_art_2013,itti2014computationalmodels} for extensive reviews of the literature.
New models are published on a regular basis with contributions coming mainly from the communities of computer vision and psychology.
It has been extensively discussed which effects are important for fixation prediction, from low and high-level influences \cite{vincent2009,einhauser_objects_2008,borji_objects_2013,cerf2007,itti2005quantifying,bruce2016deeperlook,kuemmerer2017iccv} to biases \cite{Tatler2007Centerbias,Tatler2005ROC,Tatler2008,bruce2015whatsleft}, tasks \cite{rothkopf2007,koehler2014,Tatler2011EyeGuidance} and semantic effects \cite{bylinskii2016wherelooknext}.
Over time, the concept of a saliency map has moved away from its origins in low-level feature integration, and can now refer more generally to ``a map that predicts fixations''.
In practice, saliency maps are now synonymous with saliency models.

The large number of models created the need for quantitative metrics to assess progress in the field and compare models.
Many different metrics have been proposed.
The AUC-type metrics \cite{Tatler2005ROC} used to be most common while the last years have seen a shift towards metrics like CC \cite{ouerhani2003empirical}, NSS \cite{Peters2005NSS} and SIM \cite{Judd2012}, and recently the information gain metric has been proposed \cite{kuemmerer2015}. For an overview of the different metrics in use see e.g.~\cite{borji_state_of_the_art_2013,Judd2012}. The community uses these metrics in benchmarks to keep track of the progress: the MIT Saliency Benchmark \cite{mit-saliency-benchmark,Judd2012} and the LSUN Challenge \cite{LSUN2017-competition,SALICON2017-challenge,xu15turkergaze,Jiang2015}.

The most widely accepted MIT benchmark evaluates submissions in eight different metrics.
Depending on which metric one chooses, the model rankings and performances change dramatically.
This fact has lead to substantial research analyzing the differences between metrics and giving recommendations in which situation to use which metric \cite{meur2013saliencymethods,wilming2011,richeSaliency2013,bylinskii2016,riche2016metrics,riche2016evaluation}.
Other authors have instead proposed new approaches to modeling and evaluation: Modeling as point processes \cite{Barthelme,schutt2017dynamicalcognitivemodels}, other loss functions \cite{jetley2016} and GLMMs \cite{nuthmann2017howwell}.

The general conclusion in the field is that the metrics measure qualitatively different things \cite{wilming2011,richeSaliency2013,bylinskii2016}, and that it is even conceptually impossible to determine a best model independent of the different metrics. 
Recently, Kümmerer et al.~\cite{kuemmerer2015} tried to argue for a unique ranking between different models by showing  that much of the disagreement between different metrics can be removed via postprocessing of the saliency maps by optimizing the saliency scale and smoothing kernel for information gain (IG, essentially log-likelihood). 

However, this does not seem to be a satisfactory solution: For one, this approach requires access to all models one wants to compare to and needs tedious postprocessing for each of them. In addition to this practical barrier the approach also suffers from the major conceptual shortcoming that optimizing for IG cannot be optimal for all metrics. In fact, we show below that the log densities proposed in \cite{kuemmerer2015} perform suboptimally on most metrics and can still produce inconsistent rankings.
Ideally one would like a model to be able to compete in all metrics on the metric's original scale with other models, even with models that are directly optimized for that metric and where only the metric performances are known.
This is not possible when evaluating on log densities as proposed in \cite{kuemmerer2015}.

\begin{figure*}[ht]
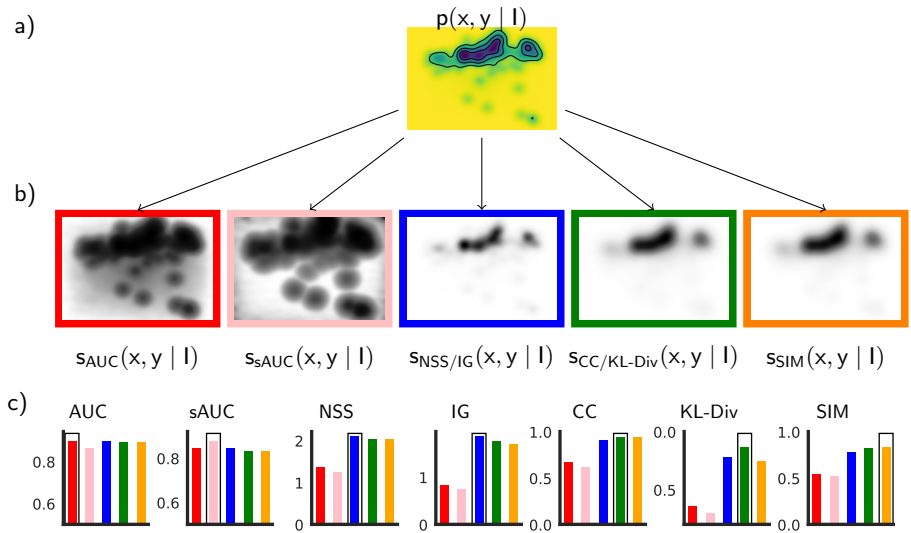

    \begin{center}
        \include{figures/introduction_figure.pgf}
    \end{center}
    \vspace{-0.6cm}
    \caption{No single saliency map can perform best in all metrics even when the true fixation distribution is known.
        This problem can be solved by separating saliency models from saliency maps.
        \textbf{a)} Fixations are distributed according to a ground truth fixation density $p(x, y \mid I)$ for some stimulus $I$ (see supplementary material for details on the visualization).
        \textbf{b)} This ground truth density predicts different saliency maps depending on the intended metric.
        The saliency maps differ dramatically due to the different properties of the metrics but always reflect the same underlying model.
        Note that the maps for the NSS and IG metrics are the same, as are those for CC and KL-Div.
        \textbf{c)} Performances of the saliency maps from b) under seven saliency metrics on a large number of fixations sampled from the model distribution in a).
        Colors of the bars correspond to the frame colors in b).
        The predicted saliency map for the specific metric (framed bar) yields best performance in all cases.
    }
    \label{fig:intro}
\end{figure*}

In fact, we show in this paper that even with knowledge of the true fixation distribution, no single saliency map can perform well in all metrics.
In practice however, researchers must still decide on a particular saliency map to submit to the benchmark.
Therefore, their model cannot compete with state-of-the-art models in all metrics -- not because the model is intrinsically bad on those metrics, but because different metrics require the saliency maps to look different, independent of the encoded information about fixation placement (see Figure~\ref{fig:intro}).
As long as one evaluates all saliency metrics on the \emph{same} saliency maps, it is impossible to solve the benchmarking problem.

Here, we argue that the fundamental problem is that saliency models and saliency maps are considered to be the same.
A major insight from Bayesian decision theory is that the derivation of optimal decisions can be decomposed into a task-independent probability distribution over possible outcomes of an experiment and a task-dependent error metric.
In the saliency setting, one decides on a saliency map to submit to a certain metric.
Correspondingly, saliency \textit{models} should be defined as metric-\textit{independent} probability densities over possible fixations and subsequently many different metric-\textit{dependent} saliency \textit{maps} can be derived from the same density for different error metrics.

We show that saliency maps for the most influential metrics AUC, sAUC, NSS, CC, SIM, and KL-Div can be derived from fixation densities in a principled way.
We demonstrate the validity of our approach on real models and real data.
By decoupling the notions of saliency models and saliency maps,
saliency models can be meaningfully compared on all metrics \textit{in their original scale},
and the MIT saliency benchmark will implement our suggested approach.

\section{Theory}


%

Motivated by the line of thoughts presented above we here propose to use the following definitions:

\begin{enumerate}
    \item a \textit{saliency model} predicts a fixation probability density $p(x, y \mid I)$ given an image $I$.
    \item a \textit{saliency metric} is a performance measure for a saliency map on ground truth data.
    \item a \textit{saliency map} $s_{p, \text{metric}}(x, y, I)$ is a metric-specific prediction derived from the model density.
\end{enumerate}

It has been argued before that formulating saliency models as probabilistic models is advantageous (e.g. \cite{Barthelme,kuemmerer2015}). In this definition, a saliency model predicts a fixation probability density, that is, the probability $p(x, y \mid I)$ of observing a fixation at a given pixel in a given image\footnote{Note that we use the fixation probability density for single fixations (as in \cite{kuemmerer2015}) whereas \cite{Barthelme} define a point process density for a whole scanpath.}.
The three definitions we propose above follow the rationale of Bayesian decision theory:
the saliency model is a posterior density over all possible events and the saliency metric is a utility function.
Based on the posterior density and the utility function, a saliency map is then chosen to maximize the expected utility.


\subsection{Predicting saliency maps from saliency models}

From the predicted fixation density of a model, one can use expected utility maximization to derive the saliency map which the model expects to yield highest performance in some metric\footnote{
    Note that the term ``metric'' is a slight abuse of notation: strictly speaking, a metric measures the distance between two objects and is usually desired to be minimal. However, in saliency, the term ``metric'' denotes the performance that one wants to maximize (with a few exceptions, e.\,g., KL-Div and earth mover's distance).}.

Evaluating a saliency metric involves a saliency map $s(x, y \mid I)$ for a stimulus $I$ and ground truth fixation data $(x_i, y_i)$.
Therefore, we can phrase a metric as a function $M[s(x, y \mid I); (x_1, y_1), \dots, (x_n, y_n)]$.
Note that some metrics as CC or SIM use an empirical saliency map instead of ground truth fixations (\textit{distribution-based metrics,richeSaliency2013}).
However, the empirical saliency map is always constructed from ground truth fixations, usually by convolving them with a Gaussian.
This can be taken to be part of the metric evaluation, as we will demonstrate below.
Simplifying notation with $D = (x_1, y_1), \dots, (x_n, y_n)$, the metric evaluation can be written as
\[
M[s(x, y \mid I); D].
\]
Assuming that the fixations are distributed according to some distribution $(x_i, y_i) \sim p(x, y \mid I)$ and therefore $D \sim \prod_1^n p(x, y)$, the expected performance of the metric on a saliency map is $\mathbb{E}_D M[s(x, y \mid I); D]$.
One should choose the saliency map which is expected to yield highest performance for the metric $M$: that is, the solution of
\[
\max_{s(x,y\mid I)}  \mathbb{E}_D M[D, s(x, y \mid I)]
\]

Solving this optimization problem for a fixation distribution $p$ given by a model of interest essentially answers the following question:
if we assume that the unknown fixations, on which the saliency map later will be evaluated, come from the model density $p$ (and therefore $D = \prod_i^n p$), what would be the best saliency map to use for metric $M$?
For a metric $M$ the solutions to the optimization problem give rise to a transformation $p(x, y \mid I) \mapsto s_M(x, y \mid I)$ from fixation densities to derived metric-specific saliency maps.
While the optimization problem might be hard in general, for most commonly-used saliency metrics it can be solved exactly or approximately, as we show below.
Importantly, the methods we outline here are deterministic transformations depending only on the model's density prediction.
No optimization using ground truth data is necessary.

In the following we give exact or approximate solutions for six of the most widely used metrics, including three metrics which operate directly on ground truth fixations (AUC, sAUC, and NSS) and three distribution-based metrics which first convert the ground truth fixations into a empirical saliency map (CC, SIM, KL-Div).
Additionally we include the IG metric introduced in \cite{kuemmerer2015} since we use this metric for converting existing saliency map models to probabilistic models.

\myparagraph{AUC, sAUC} The AUC-type metrics (``Area Under the Curve'', \cite{Tatler2005ROC}) measure the model performance in a 2AFC (2 alternative forced choice) task where the model has to decide which one of two locations has been fixated: in a 2AFC task, a system is presented with one signal and one noise stimulus and chooses which stimulus is the ``signal''. 
In the case of the AUC in saliency, signal and noise correspond to fixated and non-fixated image locations respectively (See supplementary material for a proof of the equivalence between the ROC curve and the 2AFC task).
Denoting the model's fixation distribution $p_\text{fix}(x, y)$, the nonfixation distribution $p_\text{nonfix}(x, y)$ (which is uniform for AUC and the image independent center bias for sAUC) and denote the two locations by $(x_1, y_1)$ resp. $(x_2, y_2)$.
The 2AFC task reduces to deciding whether these points are sampled from $p_\text{fix} \times p_\text{nonfix}$ or from $p_\text{nonfix} \times p_\text{fix}$.
The likelihoods of the two points given these two distributions are $p_\text{fix}(x_1, y_1) p_\text{nonfix}(x_2, y_2)$ resp. $p_\text{nonfix}(x_1, y_1) p_\text{fix}(x_2, y_2)$.
The model expects optimal performance by choosing the distribution which has higher likelihood, or equivalently, the point for which $p_\text{fix}(x, y)/p_\text{nonfix}(x, y)$ has the higher value.
Therefore the model should expect the saliency map $p_\text{fix}(x, y)/p_\text{nonfix}(x, y)$ to yield highest performance.
In the special case of the standard AUC metric, $p_\nonfix$ is constant and the saliency map boils down to $p_\fix$.
An additional practical consideration is that the MIT benchmark currently only accepts submissions as JPEG images.
To compensate for this limited precision and possible JPEG-artefacts, one should additionally histogram-equalize the saliency map (see Supplementary Material).

\myparagraph{NSS} The \textit{Normalized Scanpath Saliency} (NSS, \cite{Peters2005NSS}) performance of a saliency map model is defined to be the average saliency value of fixated pixels in the normalized (zero mean, unit variance) saliency maps (i.e., the average z-score of the fixated saliency values).

We can show analytically that one should expect the highest NSS score from the predicted fixation density itself:
given an image with $N$ pixels let the probability for a single fixation falling onto pixel $i$ be $p_i$.
Then the expected NSS of a saliency map $q=(q_1, \dots, q_N)$ with $\tfrac1N \sum_i q_i = \bar q = 0$, $\| q \|_2^2 = 1$ is $\sum_i^N p_i \cdot q_i = \langle p, q\rangle$.
Finding the saliency map with the best possible NSS is equivalent to finding the solution of the problem
\[
\max \langle p, q\rangle\quad s.t.\ \bar q= 0, \|q\|^2=1
\]

Since $q \mapsto q' = \bar p + \alpha q$ with $\alpha = \sqrt{\|p\|^2 - 1/N}$ induces a maximum-preserving bijection between $\{q \mid \bar q=0, \| q\|^2 = 1\}$ and $\{q' \mid \bar q' =  \bar p = 1/N, \|q\|^2 = \|p\|^2\}$, we can look for the maximum of $\langle p, q'\rangle\quad s.t.\ \bar q'=\bar p, \|q'\|^2=\|p\|^2$
instead 
(and normalize $q$ afterwards to get the normalized saliency map).
Because of $\langle x, y \rangle = \tfrac12 (\|x\|^2 + \|y\|^2 | - \|x-y\|^2)$, the maximum under these conditions is identical with the minimum of $\|p - q\|^2$, which is $p$.

Therefore, the best possible saliency map with respect to NSS is the density of the fixation distribution.

\myparagraph{IG} The \textit{information gain} (IG, \cite{kuemmerer2015}) metric requires the saliency map to be a probability distribution and compares the average log-probability of fixated pixels to that given by a baseline model (usually the centerbias or a uniform model).
The optimal saliency map for IG depends on how the metric interprets saliency maps as probability densities.
We normalize the saliency maps to be probability vectors (nonnegative, unit sum)
and in this case the predicted density itself yields the highest expected performance:
Let $p = (p_1, \dots, p_N)$ with $p \ge 0$, $\sum_i p_i = 1$ denote the predicted probabilities for each pixel and $q$ with $q \ge 0$, $\sum_i q_i = 1$ a saliency map. Let $p_{bl} = (p_{bl,1}, \dots, p_{bl,N})$ be the pixel probabilities of the baseline model. Then the expected IG of $q$ is
$\EE_p IG(q) = \sum_i p_i(\log q_i - \log p_{bl,i})$ and its maximum is $\argmax_q \EE_p IG(q) = \argmax_q \sum_i p_i (\log q_i - \log p_{bl, i}) = \argmax_q \sum_i p_i \log q_i = \argmax_q \sum_i p_i (\log q_i - \log p_i) = \argmin_q \sum_i p_i(\log p_i - \log q_i) = \argmin_q KL[p, q] = p$.

\myparagraph{CC} The\textit{ correlation coefficient} (CC, \cite{ouerhani2003empirical}) measures the correlation between model saliency map and empirical saliency map after normalizing both saliency maps to have zero mean and unit variance.
This is equivalent to measuring the euclidean distance between the predicted saliency map and the normalized empirical saliency map.
The expected euclidean distance to a random variable is minimized by its expectation value.
Therefore the optimal saliency map with respect to CC is the expected normalized empirical saliency map.


This shows that predicting the optimal saliency map for CC crucially depends on how the empirical saliency maps are computed.
Empirical saliency maps are typically computed by blurring observed fixation positions from eye movement data with a Gaussian kernel of a certain size.
In this case the expected empirical saliency map would be
$\EE_{x_i \sim p} \frac{1}{N} \sum_i^N G_\sigma(x)
= \frac{1}{N}\sum_i^N \EE_{x \sim p} G_\sigma(x)
= \frac{1}{N} \sum_i^N G_\sigma * p
= G_\sigma * p,
$
that is, the density blurred with a Gaussian kernel of size $\sigma$.


Unfortunately, the expected empirical saliency map is not the expected normalized empirical saliency map which was earlier shown to be optimal for CC.
Normalization involves subtracting the mean and dividing by the standard deviation, and the latter is nonlinear.
Effectively, normalizing the variance just changes the weight by which the different empirical saliency maps are averaged in the expectation value.
As long as the variances of the different empirical saliency maps don't differ too much, this won't have much of an effect and our simulations suggest that this is the case (Supplementary Material).
Therefore, as an approximation to the expected normalized empirical saliency map, we use the expected saliency map in this paper, which is computed by convolving the expected density by a Gaussian.

Obviously, if more involved techniques are used to compute the empirical saliency maps (e.g. cross validation of the kernel size as in \cite{kuemmerer2015}), then the expected empirical saliency map is harder or impossible to calculate analytically.
However, one can still approximate it numerically by sampling normalized empirical saliency maps from the expected fixation distribution and averaging them.

\myparagraph{KL-Div} The KL-Div metric computes the \textit{Kullback-Leibler divergence} between the empirical saliency maps and the model saliency maps after converting both of them into probability distributions (by making them nonnegative and normalizing them to have unit sum)
Therefore, unlike for most other metrics, in KL-Div lower values are better.

We can show that for the KL-Div metric, the expected empirical saliency map expects the best performance:
let $e = (e_1, \dots, e_N)$ with $e \ge 0$, $\sum_i e_i = 1$ denote the random variable which represents the empirical saliency map and $q$ with $q \ge 0$, $\sum_i q_i = 1$ the model saliency map.
Then we are looking for the $q$ which minimizes $\EE_p KL[e, q]$. Since
$\EE_p[KL[e, q]] = \EE_p \left[ \sum_i e_i \frac{\log e_i}{\log q_i}\right]
= \EE_p\left[ \sum_i e_i \log e_i \right] - \sum_i \EE_p[e_i] \log q_i,
$
this is equivalent to finding the maximum of $\sum_i \EE_p[e_i] \log q_i$, which is again equivalent to finding the minimum of \linebreak $\sum_i \EE_p[e_i] \log \EE_p[e_i] -\allowbreak \sum_i \EE_p[e_i] \log q_i = KL[\EE_p[e], q]$.
This is obviously minimized by $q=\EE_p[e]$, the expected empirical saliency map.
As for CC, this is the density blurred by the same kernel size as used for the empirical saliency map.

\myparagraph{SIM} The \textit{Similarity} (SIM, \cite{Judd2012}) metric normalizes the model saliency map and the empirical saliency map to be probability vectors (in the same way as KL-Div) and sums the pixelwise minimum of two saliency maps.
As opposed to the CC-metric, which can be interpreted as measuring the $l_2$-distance between normalized saliency maps, this effectively measures the $l_1$-distance between saliency maps ($
\sum_i \min(p_i, q_i) = \sum_i \frac12 \left(p_i + q_i - |p_i - q_i| \right) 
= 1 - \frac12 \| p - q \|_1,
$)
This optimization problem cannot be solved analytically in general.
Instead we solve it numerically: we perform a constrained stochastic gradient descend on sets of fixations sampled from the probability density (see Section \ref{sec:methods_predicting_saliency_maps} for details).
Note that the optimal saliency map for SIM, unlike all other saliency maps presented here, depends on the number of fixations per image (see the Supplement for details on this effect).

\section{Experiments and Results}
\label{sec:methods_predicting_saliency_maps}

We use the pysaliency toolbox \cite{pysaliency} to compute saliency metrics (see Supplement for details).
From a probability density over an image we compute five types of saliency maps:
\textbf{AUC saliency maps} are created by equalizing the probability density to yield a uniform histogram over all pixels.
\textbf{sAUC saliency maps} are created by dividing the probability density by the center bias density and again equalizing the saliency map to yield a uniform histogram over all pixels. The center bias density was estimated using a Gaussian kernel density estimate over all fixations from the MIT1003 dataset and crossvalidated across images.
\textbf{NSS/IG saliency maps} are simply the probability density.
\textbf{CC/KL-Div saliency maps} are calculated by convolving the probability density with a Gaussian kernel with $\sigma=35px$ (corresponding to 1dva, as commonly used on the MIT1003 dataset).
\textbf{SIM saliency maps}: We divide the CC saliency map by its sum to normalize it.
Starting from there, we perform constrained (nonnegative, unit sum) stochastic gradient descend on fixations sampled from the predicted density to maximize the expected SIM performance (see Supplementary Material for implementation details).

\subsection{No saliency map to rule them all}

Here we illustrate using simulated data that even if the true fixation density is known, no single saliency map can win in all saliency metrics.
From a fictional fixation density (Figure \ref{fig:intro}a) we compute the saliency maps that we predict to be optimal for the seven saliency metrics AUC, sAUC, NSS/IG, CC/KL-Div and SIM (Figure \ref{fig:intro}b).
We sample 1000 sets of 100 fixations from the fixation density and evaluate all five saliency maps using the seven different saliency metrics on this dataset (Figure \ref{fig:intro}c, raw data in the Supplement).

Although the saliency maps in Figure \ref{fig:intro}b all are predicted by the same model, they appear visually different:
while the AUC saliency map is essentially just the normalized density, the sAUC saliency map removes the center bias contribution (see above).
The NSS/IG saliency map is exactly the density and shows large areas with very low values.
The CC/KL-Div saliency map, being a blurred version of the density, is much smoother than the NSS saliency map.
The SIM saliency map looks mostly like the CC/KL-Div saliency map but is slightly more sparse.

The ranking of the five saliency maps is highly inconsistent across metrics (Figure \ref{fig:intro}c):
even with knowledge of the real fixation distribution, no saliency map can be optimal for all saliency metrics.
However, each saliency map is optimal for exactly those metrics for which it has been predicted to be optimal (framed bars).
This illustrates our main result:
By deriving metric-specific saliency maps in a principled way from fixation densities, one model can perform optimally in all metrics.
Notice that in current practice, the situation faced by an individual research team is rather to pick from one of the maps in Figure \ref{fig:intro}b and be penalized accordingly on other metrics in Figure \ref{fig:intro}c.

%

\begin{figure*}[ht!]
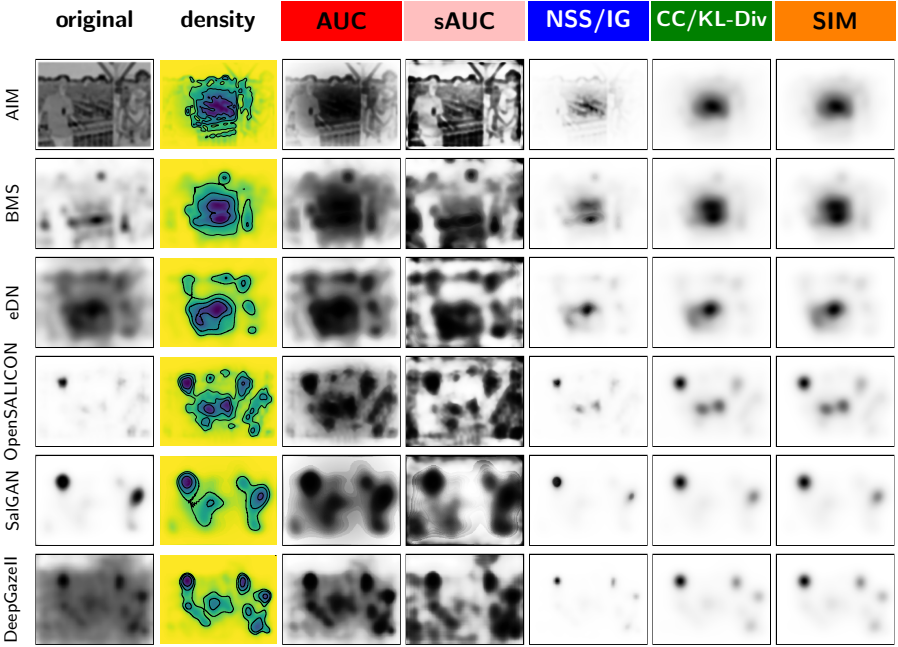

    \begin{center}
        \include{figures/example_saliency_maps.pgf}
    \end{center}
    \vspace{-0.6cm}
    \caption{The predicted saliency map for various metrics according to different models, for the same stimulus.
        For six models (rows) we show their original saliency map (first column), the probability distribution after converting the model into a probabilistic model (second column) and the saliency maps predicted for seven different metrics (columns three through seven). 
        The predictions of different models for the same metric (column) appear more similar than the predictions of the same model for different metrics (row).
        In particular, note the inconsistency of the original models (what are typically compared on the benchmark) relative to the per-metric saliency maps. 
        It is therefore difficult to visually compare original model predictions, which have been formulated for different metrics.
    }
    \label{fig:saliency_maps}
\end{figure*}

\subsection{MIT1003}

In our main experiment, we use our approach to evaluate six saliency models on the popular benchmarking dataset MIT1003 (freeviewing fixations of 15 subjects on 1003 images, \cite{Judd2009Model}).
For all evaluated models, the original source code and default parameters have been used.
The included models are
\textbf{AIM} \cite{BruceTsotso2009Saliency},
Boolean Map-based Saliency \textbf{(BMS)} \cite{zhang2013saliency},
the Ensemble of Deep Networks \textbf{(eDN)} \cite{Vig2014},
\textbf{OpenSALICON} \cite{Thomas2016},
\textbf{SalGAN} \cite{pan2017salgan}
and \textbf{DeepGaze II} \cite{kuemmerer2017iccv}.

Converting existing models that produce arbitrary saliency maps into probabilistic models is not straightforward \cite{kuemmerer2015}.
We used the method described in \cite{kuemmerer2015} and implemented in the pysaliency toolbox as \texttt{optimize\_for\_information\_gain}:
we fitted a pixelwise monotone nonlinearity and a center bias for each model to yield maximum information gain for the MIT1003 dataset (see supplementary material for details).
Unlike \cite{kuemmerer2015} we did not optimize an additional Gaussian convolution to smooth the predictions.
Since DeepGaze II is already formulated as a probabilistic model, there was no need to convert this model. For showing the ``original saliency map'' we use the log density in this case.

\myparagraph{Example saliency maps.} 

In Figure \ref{fig:saliency_maps}, we show the probability distribution and the predicted saliency maps (columns) for the saliency models (rows) for one example stimulus.
Comparing the saliency maps within and between columns, i.e.~metrics, one notices that the process of predicting saliency maps for certain metrics has a strong effect on the shape of the saliency maps that is consistent across models.
It influences the visual appearance of the saliency map to a larger degree than the actual model does:
the AUC and sAUC maps are very high contrast, while the NSS and CC saliency maps have large areas of very little saliency.
The CC and SIM saliency maps are much smoother than all other saliency maps.
It is a quite common technique in the field to compare the saliency maps of different models visually (e.g., see \cite{cornia_predicting_2016}, Figure 6; \cite{borji_quantitative_2013}, Figure 6; \cite{borji_state_of_the_art_2013}, Figure 9).
Figure \ref{fig:saliency_maps} shows that this technique can be very misleading unless the saliency maps are of the same type (i.e.~intended for the same saliency metric).

\begin{figure*}[ht!]
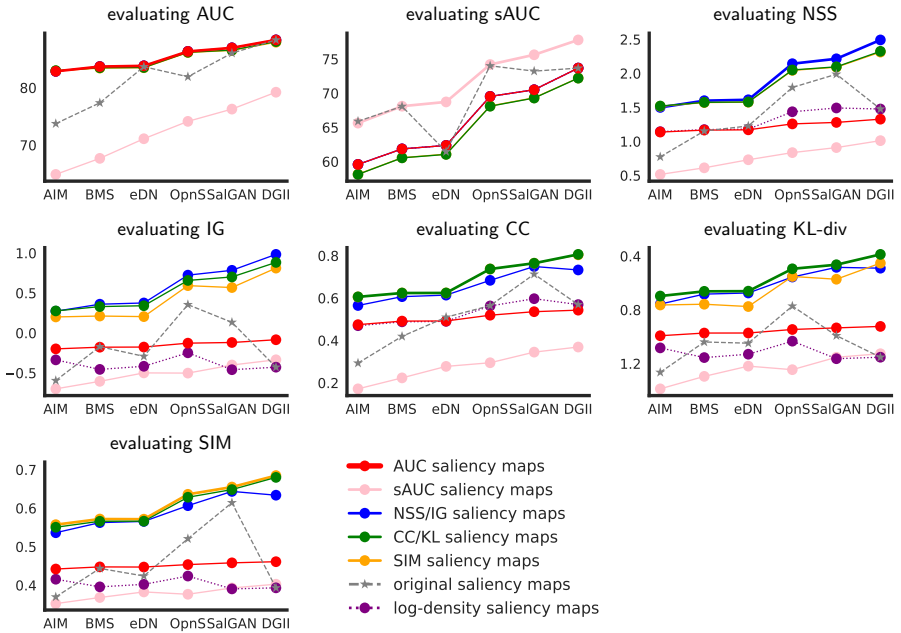

    \begin{center}
        \include{figures/evaluation_figure_small.pgf}
    \end{center}
    \vspace{-0.8cm}
    \caption{We reformulated several saliency models in terms of fixation densities and evaluated AUC, sAUC, NSS, IG, CC, KL-Div and SIM on the original saliency maps (dashed line) and the saliency maps derived from the probabilistic model for the different saliency metrics (solid lines) on the MIT1003 dataset. 
        Saliency maps derived for a given metric always yield the highest performance for that metric(thick line), and for each metric the model ranking is consistent when using the correct saliency maps -- unlike for the original saliency maps and some other derived saliency maps. 
        Note that AUC metrics yield identical results on AUC saliency maps, NSS saliency maps and log-density saliency maps, therefore the blue and purple lines are hidden by the red line in the AUC and sAUC plots. Also, the CC metric yields only slightly worse results on the SIM saliency map than on the CC saliency map, therefore the orange line is hidden by the green line in the CC plot.
        OpnS=OpenSALICON, DGII=DeepGaze II.
    }
    \label{fig:main_figure}
\end{figure*}

\myparagraph{Comparing model performance.}
In Figure \ref{fig:main_figure} we evaluate the saliency maps of the saliency models (AIM, BMS, eDN, OpenSALICON, SalGAN, DeepGaze II; x-axis) on the seven saliency metrics (subplots, raw data in the Supplement).
Each line indicates the models' performances in the evaluated metric when using a specific type of saliency map.
The dashed lines indicate performance using the models' original saliency maps (i.e.~not transformed into true probability densities).
The performances are very inconsistent between the different metrics on the original saliency maps.
The solid lines indicate the metric performances on the five types of derived saliency maps (red: AUC, pink: sAUC, blue: NSS and IG, green: CC and KL-Div, orange: SIM).
Additionally, we included log-density saliency maps as proposed in \cite{kuemmerer2015} (purple dotted lines).

For each metric, the saliency map predicted for that metric (thick line in each sub plot) yields highest performance for all models.
Conversely, saliency maps derived for other metrics often incur severe penalties (except for very few borderline cases, see below).
While the model rankings given by the different metrics on each saliency map type are much more consistent than on the original saliency maps, there is still disagreement between metrics left when evaluating all metrics on the same saliency map type.

Interestingly, the AIM model reaches better NSS performance with the CC saliency map than with the NSS saliency map.
This is easy to explain:
the AIM model's predicted density improves after blurring.
For the better models this effect vanishes. 
For example, DeepGaze II reaches significantly higher NSS scores with the NSS saliency map than with the CC saliency map and vice versa for the CC metric.
The SIM metric seems to show only slighly better performance on the SIM saliency map than on the CC saliency map, with the average difference being just 0.006.
However, the best five models with respect to SIM in the MIT Saliency Benchmark perform within a range of less than 0.02.
A difference of 0.006 could easily change a model's ranking by multiple places.

Figure \ref{fig:main_figure} also serves to illustrate a key difference between the metric unification proposed in \cite{kuemmerer2015} and our method of predicting saliency maps from fixation densities:
the metric results presented in \cite{kuemmerer2015} correspond to the purple dotted log-density lines for AUC, sAUC, NSS and to the blue density lines for IG and KL-Div (in our implementation taking the logarithm of the density is part of the metric itself).
As reported in \cite{kuemmerer2015}, the model rankings are more consistent for those lines than for the original saliency maps.
However, except for AUC and IG, in all other metrics the models are penalized when evaluated like this and additionally for the best models even the agreement between metric rankings is lost (SalGAN vs DeepGaze II, AUC/sAUC/IG vs NSS/CC/KL-Div).
This shows that the method proposed in \cite{kuemmerer2015}, while managing to remove a significant amount of the disagreement between metrics, is not perfect.

To summarize, Figure \ref{fig:main_figure} illustrates the main result of this paper:
No matter what saliency map type you decide for, even state-of-the-art models will perform suboptimally in some metrics and rankings will still be inconsistent.
Only by using the right saliency map for each metric given the model density, every model performs as well as it can theoretically and all model rankings agree.
Consequently, our evaluation yields a unique winner of the benchmark: from all included models, DeepGaze II performs best in all considered metrics.

\section{Discussion}

Despite much progress in fixation prediction in recent years, comparing saliency models to each other can be confusing due to the large number of benchmarking metrics, giving inconsistent model rankings.
Here we argue that benchmarking can be simplified by considering \textit{saliency models} to be probability density predictors, 
\textit{saliency metrics} to be performance measures that assess saliency maps against ground truth fixations, and subsequently \textit{saliency maps} to be metric-specific predictions derived from the model's density.
We have shown that probabilistic models can predict good saliency maps for the most common saliency metrics: ``good models'' perform well in many metrics.

Importantly, this metric-specific prediction reflects the same underlying model.
It is not the case that the model is being re-trained for each metric. 
Rather, the saliency maps we show are derived deterministically from the fixation density predicted by a model.
In this way it is possible to obtain optimal predictions from a given saliency density for arbitrary metrics without retraining.
The saliency model density captures all necessary information in the training data and represents it in a way that it can readily be used in combination with arbitrary error metrics.
Information gain (equivalently, log-likelihood) is an ideal optimization metric because it reflects all information in the structure of the fixation density, independent of any particular metric.
Therefore, it should lead to good results in all metrics.

The fact that metrics impose strong constraints on saliency maps means that it is misleading to visually compare saliency maps intended for different metrics (see Figure \ref{fig:saliency_maps})---but this is commonly done in the field (\cite{cornia_predicting_2016,borji_quantitative_2013,borji_state_of_the_art_2013})
For example, the optimal saliency maps for distribution-based metrics like CC, SIM and KL-Div require blurring unlike those for NSS and IG.

Another consequence of the present work is that the eight metrics available on the MIT benchmark can now be seen as a benefit rather than a possible source of confusion.
Since each metric assesses different aspects of the fixation prediction, the benchmark would now allow fair comparison over a number of tasks of interest, which may be more or less relevant for certain applications.
For example, sAUC is most relevant when one is interested in a model's predictive performance once the center bias is excluded (e.g., in applying to a setting with a different center bias from the MIT1003 training data).

While the saliency maps we have derived give the optimal metric-specific saliency map for a given fixation density, it is nevertheless still possible that a given model could do better on a metric with a saliency map not intended for that metric, rather than the metric-specific saliency map itself.
If the model's density is not the correct one (i.e. does not reflect the data-generating density), then the derived saliency maps can be suboptimal.
If the model's density is especially bad, some metrics might even perform better on saliency maps not predicted for this metric than on the one predicted for this metric.
For example: if a model's density prediction is too sparse, the AUC metric will perform better on the smoothed CC saliency map than it will perform on the actual AUC saliency map.
Therefore, actually optimizing model predictions for each specific metric may yield insights into the differences between the metrics (by comparing the underlying densities).
Indeed, this could in practice produce better performance on the training metric than an information gain optimized density.
The fact that we don't observe this effect on the original saliency maps (which \textit{were} trained in the case of eDN, OpenSALICON, SalGAN and DeepGaze II: Figure \ref{fig:main_figure}, dashed lines) suggests any improvement is likely small, and can come at the price of performing substantially worse in other metrics.

Finally, we would like to note that the distinction between saliency models and saliency maps we draw here does not contradict ideas that a ``saliency map'' or maps may be instantiated in the human brain, as a corollary of bottom-up attentional guidance or an importance map for (e.g.) choosing the next place to fixate in a scene \cite{li2002saliency,treisman1980feature,koch_shifts_1985}.
Our nomenclature is rather independent and intended for saliency model benchmarking.

The code for evaluating saliency models as demonstrated in this work has been released as part of the \texttt{pysaliency} python library (available at \url{https://github.com/matthias-k/pysaliency}).


\myparagraph{Conclusion}
Our work solves the problem that one saliency model cannot reach state-of-the-art performance in all relevant saliency metrics.
Our key theoretical contribution is to decouple the notions of saliency models and saliency maps.
For benchmarking practice, this means that saliency models can be meaningfully compared on all metrics \textit{in their original scale}.
Therefore, our method allows comparing to traditional models that do not use this method; it works even if only metric scores of other models are known (as for example in cases where metric scores are published in a paper).
Practically, this means that there is no need to revise an existing benchmark: researchers who submit model densities can have their performance fairly evaluated, but existing models can remain in the table.
The MIT saliency benchmark will implement this option.




\myparagraph{Acknowledgements} This study is part of Matthias Kümmerer's thesis work at the International Max Planck Research School for Intelligent Systems (IMPRS-IS). The research has been funded by the German Science Foundation (DFG; Collaborative Research Centre 1233) and the German Excellency Initiative (EXC307).

{\small
\bibliographystyle{splncs04}
\bibliography{literature}
}

\newpage


\section{Supplementary Material}

\setcounter{figure}{3}

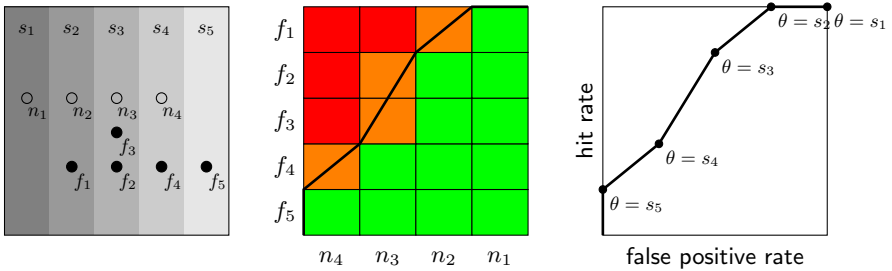
\begin{figure*}[t]
    \begin{center}
        \begin{tikzpicture}[font=\small\sffamily]

  \usetikzlibrary{calc}
  \usetikzlibrary{math}


  \coordinate (image_origin) at (0, 0);
  \coordinate (image_width) at (3, 0);
  \coordinate (image_height) at (0, 3);
  \coordinate (fixlabelsep) at (0.15, -0.2);
  \coordinate (image_saliency_label_sep) at ($ 0.1*(image_width) + 0.9*(image_height) $);


  \coordinate (grid_origin) at (4, 0);
  \coordinate (grid_width) at (3,0);
  \coordinate (grid_height) at (0,3);
  \coordinate (grid_xlabel_sep) at (0, -0.3);
  \coordinate (grid_ylabel_sep) at (-0.25, 0);


  \coordinate (auc_origin) at (8, 0);
  \coordinate (auc_width) at (3,0);
  \coordinate (auc_height) at (0,3);
  \coordinate (auc_sval_label_sep) at (0.0, -0.0);
  \coordinate (auc_xlabel_sep) at (0, -0.3);
  \coordinate (auc_ylabel_sep) at (-0.25, 0);

  \tikzset{fixation/.style={radius=0.07cm,fill=black}};
  \tikzset{nonfixation/.style={radius=0.07cm}};
  \tikzset{fixlabel/.style={scale=0.8}};
  \tikzset{sallabel/.style={scale=0.8}};
  \tikzset{fpbox/.style={fill=red}};
  \tikzset{eqbox/.style={fill=orange}};
  \tikzset{cnbox/.style={fill=green}};
  \tikzset{curve/.style={line width=1pt}};

  \tikzset{aucpoint/.style={radius=0.05cm,fill=black}};
  \tikzset{auclabel/.style={scale=0.8}};

  \coordinate (f1) at ($ 0.3*(image_width) + 0.3*(image_height) $);
  \coordinate (f2) at ($ 0.5*(image_width) + 0.3*(image_height) $);
  \coordinate (f3) at ($ 0.5*(image_width) + 0.45*(image_height) $);
  \coordinate (f4) at ($ 0.7*(image_width) + 0.3*(image_height) $);
  \coordinate (f5) at ($ 0.9*(image_width) + 0.3*(image_height) $);

  \coordinate (n1) at ($ 0.1*(image_width) + 0.6*(image_height) $);
  \coordinate (n2) at ($ 0.3*(image_width) + 0.6*(image_height) $);
  \coordinate (n3) at ($ 0.5*(image_width) + 0.6*(image_height) $);
  \coordinate (n4) at ($ 0.7*(image_width) + 0.6*(image_height) $);

  \definecolor{fillcolor5}{rgb}{0.9, 0.9, 0.9};
  \definecolor{fillcolor4}{rgb}{0.8, 0.8, 0.8};
  \definecolor{fillcolor3}{rgb}{0.7, 0.7, 0.7};
  \definecolor{fillcolor2}{rgb}{0.6, 0.6, 0.6};
  \definecolor{fillcolor1}{rgb}{0.5, 0.5, 0.5};
  \fill[fill=fillcolor1] ($ (image_origin) + 0.0*(image_width) $) rectangle ++($ 0.2*(image_width) + (image_height) $);
  \fill[fill=fillcolor2] ($ (image_origin) + 0.2*(image_width) $) rectangle ++($ 0.2*(image_width) + (image_height) $);
  \fill[fill=fillcolor3] ($ (image_origin) + 0.4*(image_width) $) rectangle ++($ 0.2*(image_width) + (image_height) $);
  \fill[fill=fillcolor4] ($ (image_origin) + 0.6*(image_width) $) rectangle ++($ 0.2*(image_width) + (image_height) $);
  \fill[fill=fillcolor5] ($ (image_origin) + 0.8*(image_width) $) rectangle ++($ 0.2*(image_width) + (image_height) $);
  \draw[-] (image_origin) rectangle ++($ (image_height) + (image_width) $);

  \foreach[evaluate={\npos using 0.2*\ncount}] \ncount in {0, 1, 2, 3, 4} {
    \tikzmath{
      \nval = int(\ncount + 1);
    }
    \node[sallabel] at ($ (image_origin) + \npos*(image_width) + (image_saliency_label_sep) $) {$s_\nval$};
  };

  \draw[fixation] (f1) circle;
  \draw[fixation] (f2) circle;
  \draw[fixation] (f3) circle;
  \draw[fixation] (f4) circle;
  \draw[fixation] (f5) circle;

  \node[fixlabel] at ($ (f1) + (fixlabelsep) $) {$f_1$};
  \node[fixlabel] at ($ (f2) + (fixlabelsep) $) {$f_2$};
  \node[fixlabel] at ($ (f3) + (fixlabelsep) $) {$f_3$};
  \node[fixlabel] at ($ (f4) + (fixlabelsep) $) {$f_4$};
  \node[fixlabel] at ($ (f5) + (fixlabelsep) $) {$f_5$};

  \draw[nonfixation] (n1) circle;
  \draw[nonfixation] (n2) circle;
  \draw[nonfixation] (n3) circle;
  \draw[nonfixation] (n4) circle;

  \node[fixlabel] at ($ (n1) + (fixlabelsep) $) {$n_1$};
  \node[fixlabel] at ($ (n2) + (fixlabelsep) $) {$n_2$};
  \node[fixlabel] at ($ (n3) + (fixlabelsep) $) {$n_3$};
  \node[fixlabel] at ($ (n4) + (fixlabelsep) $) {$n_4$};



  \draw[cnbox] ($ (grid_origin) + 0.0*(grid_height) + 0.0*(grid_width) $) rectangle ++($ 0.25*(grid_width) + 0.2*(grid_height) $);
  \draw[cnbox] ($ (grid_origin) + 0.0*(grid_height) + 0.25*(grid_width) $) rectangle ++($ 0.25*(grid_width) + 0.2*(grid_height) $);
  \draw[cnbox] ($ (grid_origin) + 0.0*(grid_height) + 0.5*(grid_width) $) rectangle ++($ 0.25*(grid_width) + 0.2*(grid_height) $);
  \draw[cnbox] ($ (grid_origin) + 0.0*(grid_height) + 0.75*(grid_width) $) rectangle ++($ 0.25*(grid_width) + 0.2*(grid_height) $);
  
  \draw[eqbox] ($ (grid_origin) + 0.2*(grid_height) + 0.0*(grid_width) $) rectangle ++($ 0.25*(grid_width) + 0.2*(grid_height) $);
  \draw[cnbox] ($ (grid_origin) + 0.2*(grid_height) + 0.25*(grid_width) $) rectangle ++($ 0.25*(grid_width) + 0.2*(grid_height) $);
  \draw[cnbox] ($ (grid_origin) + 0.2*(grid_height) + 0.5*(grid_width) $) rectangle ++($ 0.25*(grid_width) + 0.2*(grid_height) $);
  \draw[cnbox] ($ (grid_origin) + 0.2*(grid_height) + 0.75*(grid_width) $) rectangle ++($ 0.25*(grid_width) + 0.2*(grid_height) $);

  \draw[fpbox] ($ (grid_origin) + 0.4*(grid_height) + 0.0*(grid_width) $) rectangle ++($ 0.25*(grid_width) + 0.2*(grid_height) $);
  \draw[eqbox] ($ (grid_origin) + 0.4*(grid_height) + 0.25*(grid_width) $) rectangle ++($ 0.25*(grid_width) + 0.2*(grid_height) $);
  \draw[cnbox] ($ (grid_origin) + 0.4*(grid_height) + 0.5*(grid_width) $) rectangle ++($ 0.25*(grid_width) + 0.2*(grid_height) $);
  \draw[cnbox] ($ (grid_origin) + 0.4*(grid_height) + 0.75*(grid_width) $) rectangle ++($ 0.25*(grid_width) + 0.2*(grid_height) $);

  \draw[fpbox] ($ (grid_origin) + 0.6*(grid_height) + 0.0*(grid_width) $) rectangle ++($ 0.25*(grid_width) + 0.2*(grid_height) $);
  \draw[eqbox] ($ (grid_origin) + 0.6*(grid_height) + 0.25*(grid_width) $) rectangle ++($ 0.25*(grid_width) + 0.2*(grid_height) $);
  \draw[cnbox] ($ (grid_origin) + 0.6*(grid_height) + 0.5*(grid_width) $) rectangle ++($ 0.25*(grid_width) + 0.2*(grid_height) $);
  \draw[cnbox] ($ (grid_origin) + 0.6*(grid_height) + 0.75*(grid_width) $) rectangle ++($ 0.25*(grid_width) + 0.2*(grid_height) $);

  \draw[fpbox] ($ (grid_origin) + 0.8*(grid_height) + 0.0*(grid_width) $) rectangle ++($ 0.25*(grid_width) + 0.2*(grid_height) $);
  \draw[fpbox] ($ (grid_origin) + 0.8*(grid_height) + 0.25*(grid_width) $) rectangle ++($ 0.25*(grid_width) + 0.2*(grid_height) $);
  \draw[eqbox] ($ (grid_origin) + 0.8*(grid_height) + 0.5*(grid_width) $) rectangle ++($ 0.25*(grid_width) + 0.2*(grid_height) $);
  \draw[cnbox] ($ (grid_origin) + 0.8*(grid_height) + 0.75*(grid_width) $) rectangle ++($ 0.25*(grid_width) + 0.2*(grid_height) $);

  \foreach[evaluate={\npos using 0.25*\ncount}] \ncount in {0, 1, 2, 3} {
    \tikzmath{
      \nval = int(4 - \ncount);
    }
    \node at ($ (grid_origin) + \npos*(grid_width) + 0.125*(grid_width) + (grid_xlabel_sep) $) {$n_\nval$};
  };

  \foreach[evaluate={\npos using 0.2*\ncount}] \ncount in {0, 1, 2, 3, 4} {
    \tikzmath{
      \nval = int(5 -\ncount );
    }
    \node at ($ (grid_origin) + \npos*(grid_height) + 0.1*(grid_height) + (grid_ylabel_sep) $) {$f_\nval$};
  };

  \draw[-,curve] (grid_origin)
            -- ++($ 0.2*(grid_height) + 0.0*(grid_width) $)
            -- ++($ 0.2*(grid_height) + 0.25*(grid_width) $)
            -- ++($ 0.4*(grid_height) + 0.25*(grid_width) $)
            -- ++($ 0.2*(grid_height) + 0.25*(grid_width) $)
            -- ++($ 0.0*(grid_height) + 0.25*(grid_width) $)
            ;


  \draw[-] (auc_origin) rectangle ++($ (auc_height) + (auc_width) $);
  \coordinate (lastp) at (auc_origin);
  \foreach[count=\count] \fpcount/\hitcount in {0/1, 1/2, 2/4, 3/5, 4/5} {
    \tikzmath{
      \fprate = 0.25*\fpcount;
      \hitrate = 0.2*\hitcount;
      \sval = int(6 - \count);
    }
    \coordinate (newp) at ($ (auc_origin) + \fprate*(auc_width) + \hitrate*(auc_height) $);
    \draw[aucpoint] (newp) circle;
    \node[anchor=north west,auclabel] at  ($ (newp) + (auc_sval_label_sep) $) {$\theta = s_{\sval}$};
    \draw[-,curve] (lastp) -- (newp);
    \coordinate (lastp) at (newp);
  };

  \node at ($ (auc_origin) + 0.5*(auc_width) + (auc_xlabel_sep) $) {false positive rate};
  \node[rotate=90] at ($ (auc_origin) + 0.5*(auc_height) + (auc_ylabel_sep) $) {hit rate};



%
%
%
%
\end{tikzpicture}
    \end{center}
    \caption{AUC metrics measure the performance of the saliency map in a 2AFC task where the saliency values of two locations are used to decide which of these two locations is a fixation and which is a nonfixation.
        \textbf{a)} An example saliency map is shown consisting of five saliency values ($s_1 < \dots < s_5$) and with five fixations ($f_1, \dots, f_5$) and four nonfixations ($n_1, \dots, n_4)$.
        \textbf{b)} The performance in the 2AFC task can be calculated by going through all fixation-nonfixation pairs $(f_i, n_j)$:
        The saliency map decides correct if the saliency value of $f_i$ is greater than $n_j$ (green), incorrect if it is smaller (red) and has chance performance if the values are equal (orange). Below the thick line are all correct predictions (green) and half of the chance cases (orange).
        \textbf{c)} The ROC curve of the saliency map with respect to the given fixations and nonfixations. For each threshold $\theta$ all values of saliency value greater or equal to $\theta$ are classified as fixations. Comparing b) and c) shows that the area under the curve in c) is exactly the performance in the 2AFC task in b).
    }
    \label{fig:supp_auc_vs_2afc}
\end{figure*}

\begin{table}
    \begin{center}
    {\footnotesize
      \begin{tabular}{l|l|rrr}
        Saliency Map & Model & \multicolumn{2}{c}{Binning} & Difference \\
        &       & None & 8bit & \\ \hline
        density &AIM & 0.82883 & 0.82855 & 0.00028\\
&BMS & 0.83712 & 0.83676 & 0.00035\\
&eDN & 0.83836 & 0.83810 & 0.00026\\
&OpenSALICON & 0.86350 & 0.86200 & 0.00150\\
&SalGAN & 0.86973 & 0.86845 & 0.00128\\
&DeepGazeII & 0.88355 & 0.87931 & 0.00424\\\hline
equalized &AIM & 0.82883 & 0.82882 & 0.00001\\
&BMS & 0.83712 & 0.83710 & 0.00002\\
&eDN & 0.83836 & 0.83834 & 0.00001\\
&OpenSALICON & 0.86350 & 0.86347 & 0.00003\\
&SalGAN & 0.86973 & 0.86970 & 0.00003\\
&DeepGazeII & 0.88355 & 0.88351 & 0.00004\\

      \end{tabular}}
    \end{center}
    \caption{AUC and low precision: While AUC metrics in theory depend only on the ranking of the saliency values and therefore are invariant to monotone transformations, this does not hold anymore when the saliency map is saved with limited precision (e.g. as 8bit PNG/JPEG as common).
        In this case, the saliency map should be rescaled to have a uniform histogram before saving.}
    \label{tab:supp_auc_low_precision}
\end{table}

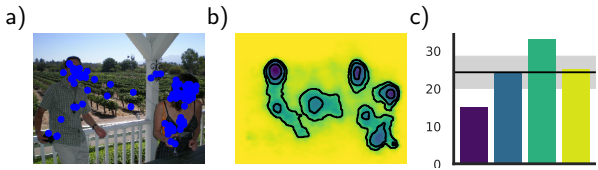
\begin{figure}
    \begin{center}
    \begin{tikzpicture}[font=\sffamily]
    \coordinate (hsep) at (2.7, 0);
    \coordinate (lsep) at (-0.08, 0.1);
    \coordinate (barlcorr) at (0.2, 0.0);
    \coordinate (barcorr) at (-0.2, 0.0);
    \tikzset{panel/.style={scale=0.85}}
    \node[panel] (stimulus) at (0, 0) {
        \includegraphics[]{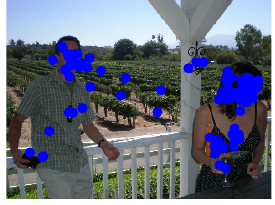}
    };
    \node[panel] (prediction) at ($ (hsep) $) {
        \includegraphics[]{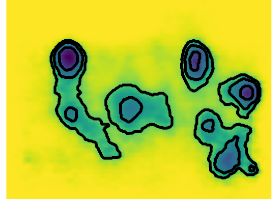}
    };
    \node[panel] (counts) at ($ 2*(hsep) + (barcorr) $) {
        \includegraphics[]{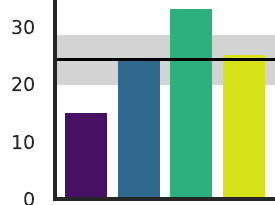}
    };
    
    \node at ($ (stimulus.north west) + (lsep) $) {{a)}};
    \node at ($ (prediction.north west) + (lsep) $) {{b)}};
    \node at ($ (counts.north west) + (lsep) + (barlcorr) $) {{c)}};
    \end{tikzpicture}
    \end{center}
    \caption{Visualizing fixation densities:
        \textbf{a)} an example stimulus with $N=97$ ground truth fixations.
        \textbf{b)} DeepGaze II predicts a fixation density for this stimulus. The contour lines separate the image into four areas of decreasing probability density such that each area has the same total probability mass.
        \textbf{c)} The number of ground truth fixations in each of the four areas. The model expects the same number of fixations for each area (horizontal line: $24.25$ fixations for $N$ fixations total).
        The gray area shows the expected standard deviation from this number.
        DeepGaze II overestimates the how peaked the density is: there are too few fixations in darkest area.
        Vice versa, it misses some probability mass in the second to last area.
        However, the large error margin (gray area) indicates that substantial deviations from the expected number of fixations are to be expected.
    }
    \label{fig:visualizing_densities}
\end{figure}

\subsection{Implementation details on the saliency metrics} We use the pysaliency toolbox \cite{pysaliency} to compute metrics.
\textbf{AUC}: We use all pixels as nonfixations. As thresholds we use the combined saliency values of all fixations and nonfixations.
\textbf{sAUC}: We use the fixations of all other images of the dataset as nonfixations. As for AUC, we use the combined saliency values of all fixations and nonfixations as thresholds.
\textbf{NSS} computes the mean saliency of fixation locations after normalizing the saliency map to have zero mean and unit variance.
\textbf{IG} computes the mean log density of fixation locations for a model's predicted fixation density and substracts the average log density of fixation locations for a baseline model's predicted fixation density.
To convert a saliency map to a probability distribution, we check whether any values of the saliency map are negative.
If so, we subtract the minimal value from the saliency to make it non-negative.
Afterwards we divide the saliency by the sum of all values.
For the baseline model we transform the coordinates of all fixations in the MIT1003 dataset to range from 0 to 1.
From these points a Gaussian kernel density estimator with a bandwidth of 0.22 is computed (the bandwidth has been tuned with leave-one-image-out crossvalidation).
The baseline model scales the density predicted by the estimator to the size of the image in question.
For images in the MIT1003 dataset (i.e. for Figure 3), only fixations from all other images in the dataset are used to compute the baseline density for the image.
\textbf{CC}: As suggested for the MIT1003 dataset used by us \cite{Judd2009Model}, we convolve the fixation maps of the ground truth fixations with a gaussian kernel with $\sigma=35px$ to compute empirical saliency maps.
\textbf{KL-Div}: We use the same empirical saliency maps as for CC and the same normalization procedure as for IG.
\textbf{SIM}: We use the same empirical saliency maps as for CC and the same normalization procedure as for IG.

\subsection{Converting saliency map models to fixation density models}

To convert existing saliency-map based models to probabilistic models, we used the method described in \cite{kuemmerer2015} and implemented in the pysaliency toolbox in the method \texttt{optimize\_for\_information\_gain}:

It first rescales all saliency maps for the dataset in question jointly to range from 0 for the smallest saliency value (over the full dataset) to 1 for the largest saliency value.
The a pixelwise montone nonlinearity is applied to each saliency map.
This nonlinearity is implemented as a continous piecewise linear function with 20 equidistant segments from 0 to 1.

The result is multiplied pixelwise with a centerbias which is parametrized with another piecewise linear function applied to
\[\sqrt{(x-\tfrac12 x_{max})^2+\alpha (y-\tfrac12y_{max})^2}/\sqrt{\tfrac14 x_{max}^2+\tfrac14\alpha y_{max}^2}
\]
 where $x_{max}$ and $y_{max}$ are the maximal $x$ and $y$ coordinates for the image in question.
The piecewise linear function for the centerbias is parametrized as a continous piecewise linear function with 12 equidistant segments from 0 to 1.
The resulting product is divided by its sum over all pixels to make it a probability distribution.

The parameters for both piecewise linear functions and the eccentricity parameter $\alpha$ are jointly optimized for maximum likelihood on the MIT1003 dataset.
Note that unlike \cite{kuemmerer2015} we did not optimize an additional Gaussian convolution to smooth the predictions.

%

\subsection{Computing saliency maps for SIM}

To compute the saliency map for the SIM metric from a model density, we first divide the CC saliency map (density convolved with a Gaussian of size 35px=1dva) by its sum to normalize it.
Starting from there, we perform a constrained stochastic gradient descend on fixations sampled from the predicted density to maximize the expected SIM performance.
The (linear) constraints that are enforced in every step of the gradient descend are nonnegativity and unit sum.
Each sample consists of 100 fixations (in correspondence to the dataset we are using).
We use a batchsize of 50 samples and start with a learning rate of $10^{-7}$.
We use a fixed set of 1000 samples as validation data.
Every 1000 training samples we compute the validation performance.
Whenever it decreases compared to the last epoch, we go back to the point of best validation performance so far and decrease the learning rate by a factor of $\tfrac13$ and continue the gradient descend.
We stop when the learning rate is smaller than $10^{-9}$.

\subsection{The AUC metrics: Digitizing saliency maps}

Digitizing the saliency map e.g. by storing them as 8bit images can obviously affect metric performance.
The AUC type metrics are sensitive only to the ranking of the saliency values and therefore especially sensitive to mapping similar saliency values to the same value.
In Table \ref{tab:supp_auc_low_precision}, we evaluate the AUC metric for all included models in four different ways:
We use either the model fixation density or we additionally transform it to have a uniform histogram.
Also, we optionally bin the saliency values to 256 different values using equidistant bins.
Since the AUC metrics are invariant to monotonic transformations, both density and equalized density should have the same AUC performance, as is indeed the case if no binning is applied.
In the case of binning, however, the performances change:
while for the normalized density binning does not affect performance a lot, for the density it does so.
The performance loss after binning the density seems to be the stronger for better models.
This is likely the case since better models will map larger areas of the image to very small values that all end up in the lowest bin.

\subsection{The CC metric: mean normalized empirical vs normalized mean empirical saliency maps}

We use the mean empirical saliency map for the CC metric.
As explained in the main text, this is an approximation:
the optimal saliency map would be the mean normalized empirical saliency map (i.e. one has to normalize the empirical saliency maps to zero mean and unit variance before taking the mean).

\begin{figure*}[t]
    \begin{center}
        \begin{tikzpicture}
        \tikzset{label/.style={font=\sffamily}};
        \coordinate (vsep) at (0.5, 0);
        \coordinate (tablesep) at (0, -0.5);
        \coordinate (labelsep) at (-0.2, -0.2);
        \node[anchor=north west] (density) at (0, 0) {
            \includegraphics[scale=0.8]{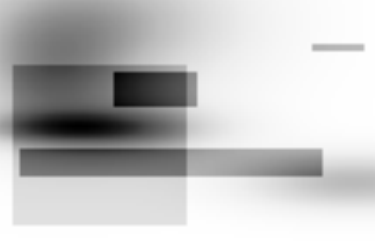}
        };
        
        \node[anchor=north west] (MES) at ($ (density.north east) + (vsep) $) {
            \includegraphics[scale=0.8]{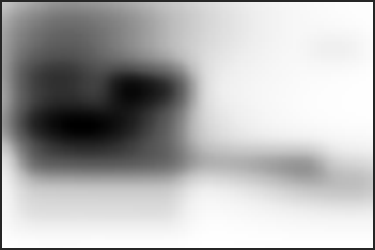}
        };
        \node[anchor=north west] (MNES) at ($ (MES.north east) + (vsep) $) {
            \includegraphics[scale=0.8]{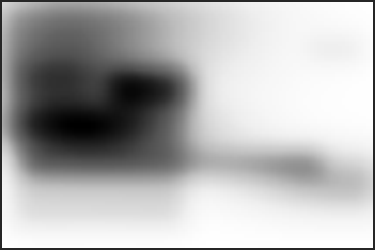}
        };
        \node[anchor=north west,scale=0.8] at ($ (density.south west) + (tablesep) $) {
\begin{tabular}{p{3cm}|c|c|c|c|c}
  \multirow{2}{*}{\diagbox[width=2.8cm]{fixations}{kernel size}}& \multicolumn{5}{c}{CC scores (mean empirical/mean normalized empirical)}\\
	& 1 & 5 & 10 & 20 & 30\\\hline
1 &0.0127/0.0127&0.0632/0.0633&0.1210/0.1214&0.2227/0.2242&0.3066/0.3100\\
10 &0.0412/0.0412&0.1928/0.1929&0.3488/0.3491&0.5630/0.5641&0.6855/0.6874\\
50 &0.0920/0.0920&0.3960/0.3960&0.6120/0.6121&0.7728/0.7731&0.8198/0.8206\\
100 &0.1295/0.1295&0.5191/0.5191&0.7307/0.7307&0.8464/0.8465&0.8707/0.8709\\
200 &0.1817/0.1817&0.6509/0.6509&0.8313/0.8314&0.9074/0.9074&0.9203/0.9204\\%
\end{tabular}

        };
        
        \node[label] at ($ (density.north west) + (labelsep) $) {\textbf{a)}};
        \node[label] at ($ (MES.north west) + (labelsep) $) {\textbf{b)}};
        \node[label] at ($ (MNES.north west) + (labelsep) $) {\textbf{c)}};
        \node[label] at ($ (density.south west) + (labelsep) + (0, -0.0) $) {\textbf{d)}};
        \end{tikzpicture}
    \end{center}
    \caption{Predicting optimal saliency maps for the CC metric:
        Starting from a density (a) we sampled 100000 sets of either 1, 10 or 100 fixations and used them to create empirical saliency maps.
        Using these empirical saliency maps, we calculated the mean empirical saliency map (shown for 10 fixations per empirical saliency map in (b)).
        Additionally, we normalized the empirical saliency maps to have zero mean and unit variance to compute the mean normalized empirical saliency map (c) which is optimal with respect to the CC metric.
        Then we sampled another 100000 empirical saliency maps from the original density and evaluated CC scores of the mean empirical and mean normalized empirical saliency maps (d).
        The mean normalized saliency map yields slighly higher scores in all cases but the difference to the mean empirical saliency map is tiny, indicating that the expected empirical saliency map is a very good approximation of the optimal saliency map for the CC metric.
    }
    \label{fig:supp_CC}
\end{figure*}
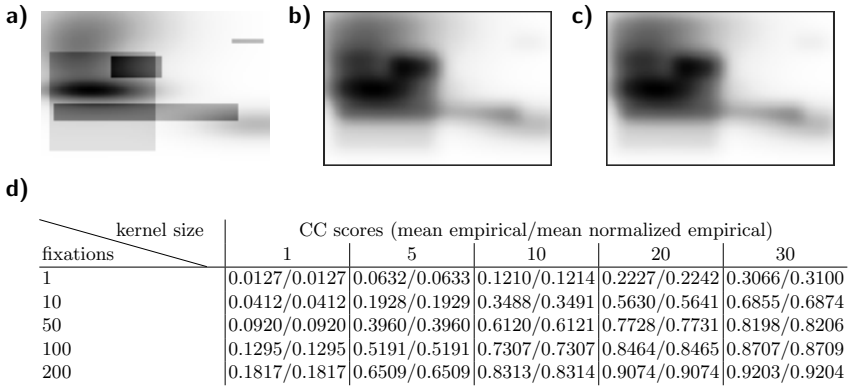

To check the validity of our approximation, we sampled fixations from a distribution (Figure \ref{fig:supp_CC}a) and used those fixations to compute average empirical saliency maps (Figure \ref{fig:supp_CC}b) and average normalized empirical saliency maps (Figure \ref{fig:supp_CC}c) for different numbers of fixations per sample and kernel sizes in the computation of empirical saliency maps.

We evaluated both types of saliency maps on newly sampled fixations and compared the CC performances (Figure \ref{fig:supp_CC}d).
The performances for both saliency maps are very close in all cases, suggesting that the mean empirical saliency map is an adequate approximation for the mean normalized saliency map when computing CC performances.

\subsection{The SIM metric depends on the number of fixations per image}

Unlike all other metrics presented in this work, the optimal saliency map for the SIM metric depends on how many fixations per image are in the dataset in question.
If ignoring the constraint that the values of the saliency map should sum up to one, this effect is easy to see:
The SIM metric effectively measures the l1 distance between empirical saliency map and model saliency map and this distance is minimized by the median empirical saliency map, which will be mostly zero if there are only very little fixations used to compute each empirical saliency map.

\begin{figure*}
    \begin{center}
        \begin{tikzpicture}
        \tikzset{label/.style={font=\sffamily,anchor=north west}};
      \node (smaps) {\includegraphics[scale=0.75]{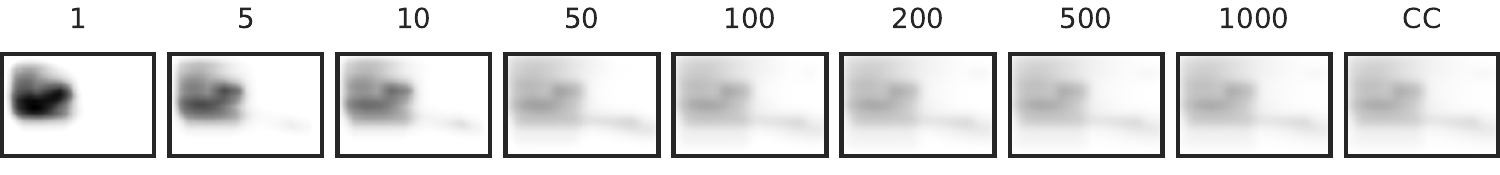}};
      
      \node[scale=0.9] (table) at ($ (smaps) + (0, 
      -4) $) {
          \begin{tabular}{l|rrrrrrrr}
          Fixations/sample & 1 & 5 & 10 & 50 & 100 & 200 & 500 & 1000 \\
          Saliency Map & \\\hline
          \input{data/SIM_fixation_dependency}
          \end{tabular}
      };
  
      \node[label] (labela) at ($ (smaps.north west) + (-0.5, 0) $) {\textbf{a)}};
     \node[label] (labelb) at (labela.north west |- table.north west) {\textbf{b)}};
    \end{tikzpicture}
    \end{center}
%
    \caption{The optimal SIM saliency map depends on the number of fixations. \textbf{(a)} For a sample density (see Figure \ref{fig:supp_CC}), we calculated the optimal SIM saliency map for different numbers of fixations per sample (numbers on top) and additionally the mean empirical saliency map (CC). \textbf{(b)} average performance of those saliency maps (rows) when repeatedly sampling a certain number of fixations (columns) and computing SIM.
    The best performing saliency map for each sampled dataset (columns) is printed in boldface.
    It's always the saliency map calculated with the same number of fixations.
    Note that the CC saliency map -- although looking identical -- always performs slighly worse}
    \label{fig:SIM-fixation-count-dependency}
\end{figure*}

This effect is still present when constraining the saliency map to have unit sum, as we demonstrate in Figure \ref{fig:SIM-fixation-count-dependency}.
For a sample density (Figure \ref{fig:supp_CC}a), we computed the optimal saliency maps for different numbers of fixations per sample according to our method detailed in Section \ref{sec:methods_predicting_saliency_maps}.
The resulting saliency maps are shown in Figure \ref{fig:SIM-fixation-count-dependency}a.
If there are only few fixations per sample, the resulting saliency maps have much larger areas of zeros, effectively being more sparse.
For more fixations per sample, the saliency maps visually converge to the CC saliency map (blurred density).

Subsequently, we evaluated those saliency maps (Figure \ref{fig:SIM-fixation-count-dependency}b, rows) on newly sampled fixations, again for different numbers of fixations per sample (Figure \ref{fig:SIM-fixation-count-dependency}b, columns).
Additionally, we included the CC saliency map.

The columns in Figure \ref{fig:SIM-fixation-count-dependency} show that the number of fixations used to compute the saliency map affects the performance:
The saliency map computed using the same number of fixations per sample always performs best (bold numbers), and all other saliency maps perform worse -- often dramatically.
Even in the case of 1000 fixations, there are still measurable differences between the saliency maps computed using 500 fixations, 1000 fixations and the CC saliency map.

\subsection{Visualizing probability densities}
\label{sec:methods_visualizing_densities}

Visualizing two dimensional densities is harder than it appears to be at the first glance:
Although the absolute density values have a very precise meaning, it is hard to read substantially more than the ranking of the values and maybe a very rough idea about the peakyness of the distribution from a color map.
When visualizing two dimensional probability densities, we add three contour lines separating the image into four areas of decreasing probability density such that each area has the same total probability mass
(i.e. the density predicts each area to receive the same number of fixations, see Figure \ref{fig:visualizing_densities}b).
If the darkest area is very small, this means the density predicts on fourth of the fixations to be clustered in a very small area.
If all areas are roughly of the same size, the density is nearly uniform.
Comparing the number of fixations in each area can serve as a simple heuristic to asses a model's quality (see Figure \ref{fig:visualizing_densities}c).

\subsection{Data}

The raw data for Figure \ref{fig:intro} can be found in Table \ref{tab:data-fig-1}, the raw data for Figure \ref{fig:main_figure} can be found in Table \ref{tab:data-fig-3}.


\begin{table*}
\begin{center}
        \begin{tabular}{l|rrrrrrr}
        Saliency Map & AUC & sAUC & NSS & IG & CC & KL-Div & SIM \\\hline
        \begin{filecontents*}{data/intro_figure_performances.csv}
saliencymap,AUC,sAUC,NSS,IG,CC,KLdiv,SIM
AUC,0.897325,0.842109,1.369418,0.826523,0.675036,0.644968,0.541042
sAUC,0.863243,0.875880,1.246049,0.741960,0.618483,0.700238,0.520718
NSS/IG,0.897325,0.842109,2.106131,1.865231,0.907441,0.221233,0.778510
CC/KL,0.892173,0.831612,2.024991,1.765046,0.939149,0.137458,0.824498
SIM,0.891833,0.831888,2.024197,1.700093,0.939007,0.253121,0.827775
\end{filecontents*}
\csvreader[late after line=\\,head to column names]{data/intro_figure_performances.csv}{}%
        {\saliencymap & \AUC & \sAUC & \NSS & \IG & \CC & \KLdiv & \SIM}%
        
    \end{tabular}
\end{center}
    \caption{The raw data plotted in Figure 1}
    \label{tab:data-fig-1}
\end{table*}

\newcommand{\performanceTable}[2]{%
    \begin{center}
        #1
    \end{center}
    \vspace{-0.65cm}
    \begin{center}
        \begin{tabular}{l|rrrrrr}
            Saliency Map & AIM & BMS & eDN & OpenSALICON & SalGAN  & DeepGaze II \\\hline
            \csvreader[late after line=\\,head to column names]{#2}{}%
            {\saliencymap & \AIM & \BMS & \eDN & \OpenSALICON & \SalGAN & \DeepGazeII}%
        \end{tabular}
    \end{center}
}

\begin{table*}
    {\relsize{-2}
     
    \performanceTable{Evaluating AUC}{data/performance_figure_performances_AUC.csv}
    \performanceTable{Evaluating sAUC}{data/performance_figure_performances_sAUC.csv}
    \performanceTable{Evaluating NSS}{data/performance_figure_performances_NSS.csv}
    \performanceTable{Evaluating IG}{data/performance_figure_performances_densityIG.csv}
    \performanceTable{Evaluating CC}{data/performance_figure_performances_CC.csv}
    \performanceTable{Evaluating KL-Div}{data/performance_figure_performances_KLdiv.csv}
    \performanceTable{Evaluating SIM}{data/performance_figure_performances_SIM.csv}
    }
    \caption{The raw data plotted in Figure 3}
    \label{tab:data-fig-3}
\end{table*}

\end{document}